\documentclass[journal, twoside]{IEEEtran}

\usepackage{times}
\usepackage{epsfig}
\usepackage{graphicx}
\usepackage{amsmath}
\usepackage{amssymb}
\usepackage{siunitx}
\usepackage{multirow}
\usepackage{multicol}
\usepackage{makecell}
\usepackage{gensymb}
\usepackage{caption}
\usepackage{colortbl}
\usepackage{subcaption}
\usepackage{textcomp}
\usepackage{siunitx}
\usepackage[dvipsnames]{xcolor} 
\usepackage[section]{placeins} 
\usepackage{threeparttable,booktabs} 
\usepackage{etoolbox}
\usepackage{tablefootnote}
\usepackage{verbatim}
\usepackage[bottom]{footmisc}
\usepackage{algorithm}
\usepackage{algorithmicx}  
\usepackage{algpseudocode}
\usepackage{diagbox}
\usepackage{stfloats}
\usepackage{times}
\usepackage[shortlabels]{enumitem}
\usepackage{graphicx}
\usepackage[notextcomp]{stix}

\appto\TPTnoteSettings{\footnotesize}

\algrenewcommand\algorithmicrequire{\textbf{Input:}}
\algrenewcommand\algorithmicensure{\textbf{Output:}}

\usepackage[pagebackref=true,breaklinks=true,letterpaper=true,colorlinks,bookmarks=false]{hyperref}

\title{Variation of Camera Parameters due to Common Physical Changes\\in Focal Length and Camera Pose}

\author{Hsin-Yi Chen, Chuan-Kai Fu and Jen-Hui Chuang\vspace*{-0.4cm}}

\begin{document}

\markboth{IEEE TRANSACTIONS ON Instrumentation and Measurement,~Vol.~xx, 2024}{CHUANG \MakeLowercase{\textit{et al.}}: VARIATION OF CAMERA PARAMETERS DUE TO COMMON PHYSICAL CHANGES IN FOCAL LENGTH AND CAMERA POSE} 

%



\maketitle

\begin{abstract}
Accurate calibration of camera intrinsic parameters is crucial to various computer vision-based applications in the fields of intelligent systems, autonomous vehicles, etc. However, existing calibration schemes are incompetent for finding general trend of the variation of camera parameters due to common physical changes. In this paper, it is demonstrated that major and minor variations due to changes in focal length and camera pose, respectively, can be identified with a recently proposed calibration method. It is readily observable from the experimental results that the former variations have different trends (directions) of principal point deviation for different types of camera, possibly due to different internal lens configurations, while the latter have very similar trends in the deviation which is most likely due to direction of gravity. Finally, to confirm the validity of such unprecedented findings, 3D to 2D reprojection errors are compared for different methods of camera calibration.
\end{abstract}


%
\IEEEpeerreviewmaketitle

\section{Introduction}\label{sec_intro}
\IEEEPARstart{A}{ccurate} calibration of camera intrinsic parameters is crucial to various computer vision-based applications, such as intelligent systems~\cite{App01_1,App01_2,App01_3} and autonomous vehicles~\cite{App02_1,App02_2,App02_3,App02_4}. Traditional camera calibration~\cite{pinhole_01,pinhole_02,3d_target_02,zhang2000}, which assumes fixed focal length, estimates intrinsic and extrinsic parameters based on pinhole camera model, where the relationship between a point in 3D world coordinate system (WCS) and its projection in 2D image coordinate system (ICS) can be established through the following steps: (i) transforming 3D WCS to align camera coordinate system (CCS) by extrinsic parameters of rotation and translation, and (ii) projecting points in CCS to 2D ICS by intrinsic parameters of principal point (PP) and focal length. However, few calibration methods consider the variation of intrinsic parameters due to common physical changes of a camera, e.g., changing camera focal length for better image quality of an object.\par

Ideally, focal length of a zoom-lens camera is changed by moving certain lenses in an afocal zoom system~\cite{zoom_action} along optical axis; therefore, optical axis, as well as PP, will remain unchanged. However, imperfect camera fabrication may result in deviated optical axis (and PP) due to focal length change. To determine the variation of camera parameters for such change, an optical experiment is established in~\cite{zoom_depend_optical}, which adopts the autocollimated laser approach to trace the trajectory of laser points shown on image in whole range of lens settings to demonstrate the variation of PP w.r.t. focal length change. Despite reasonable description of the above variation can be obtained, e.g., up to 6-pixel shifts of PP along a fixed direction as zoom motor setting changed from 100 to 11000, such approach can only be practiced in the laboratory due to the need to access unpackaged lenses.\par  

In~\cite{zoom_depend_01,zoom_depend_02,zoom_depend_03,zoom_depend_04,zoom_depend_05,zoom_depend_06,zoom_depend_07, zoom_depend_08, gravity_02, gravity_01}, several attempts are made to obtain similar variations of intrinsic parameters for commercial (packaged) zoom-lens camera, wherein some traditional camera calibration schemes are applied to a camera with different settings of focal length, before calibrated parameters are collected, and somehow fitted with some elementary hyper-surfaces. However, as descriptions of variation of camera parameters have different trends for different methods, none of them are indeed conclusive. Overall, two major factors need to be further investigated before the foregoing problem can be resolved, which include: (i) correctness of the adopted camera calibration scheme and (ii) rationality in the description of such variation.\par 

For (i), a geometry-based camera calibration method~\cite{geometry_01, geometry_02}, which is shown to be more accurate than other calibration schemes for synthetic data, is adopted in this paper, with a new evaluation scheme developed to demonstrate the correctness of such calibration method for real data. In addition, under the assumption that an internal lens system would be slightly shifted toward the direction of gravity, a novel experimental setup is designed to investigate typically small (minor) variations of camera parameters due to camera pose change, on top of the supposedly more significant variations due to changes of camera focal length, so as to address the rationality issue in (ii).\par  

If the foregoing variations can be correctly obtained and reasonably explained, adverse effects of incorrect calibration on various computer vision applications involving common physical changes of a camera may be alleviated. Specifically, according to our investigation for a number of commercial zoom-lens cameras, about 70 to 200-pixel shift of camera PP can be identified for relatively large focal length change, while about 10 to 20-pixel shift can be identified for seemly less influential camera pose change. In summary, contributions of this paper include:
\begin{enumerate}[(i)]
    \item Demonstrate major (unidirectional and monotonic) shifts in PP due to focal length change with a recently proposed calibration method.
    \item Demonstrate minor (mostly sideway and consistent to the direction of gravity) shifts in PP due to camera pose change, based on a novel experimental setup.
    \item Offer reasonable explanations of PP shifts (i) and (ii):
        \begin{enumerate}[a.]
        \item{shifts in (i) may be due to contact force from misaligned internal lens system and have different trends (in direction/magnitude) for different cameras.}
        \item{shifts in (ii) are due to non-contact force, i.e., gravity, and have similar trends for different cameras.}
        \end{enumerate}
    \item Proposed an effective quantitative evaluations through cross-validation to demonstrate
        \begin{enumerate}[a.]
        \item{the PP displacements will result in near the same amount of reprojection errors, which are unacceptable that re-calibration for each focal length change or pose change may be needed.}
        \item{similar amount of reprojection error may be obtained with no chage in focal length and camera pose but with less calibration patterns for Zhang's method, while Chuang's method achieves significantly smaller and acceptable errors under the same conditions.}
        \end{enumerate}
\end{enumerate}

\section{Related Work}
In this section, various camera calibration schemes are firstly reviewed. Some of them will then serve primary roles in the second part of the review of camera calibration works for zoom-lens camera.

\subsection{Camera Calibration Schemes}
\label{sec_2.1}
\noindent \textbf{Traditional Camera Calibration.} Traditional camera calibration can be roughly divided into (1) photogrammetric approach~\cite{3d_target_02,pinhole_02,zhang1998,zhang2000} and (2) self-calibration approach~\cite{self_calibration_01,self_calibration_02,self_calibration_03,self_calibration_04}. For (1), a calibrated target which provides sufficient measurements of 3D points is needed in the establishment of 2D-3D correspondences for solving camera parameters. As for (2), feature points from different images of a scene, captured by several cameras or camera motions, are used in the calibration without using a calibrated target. Nonetheless, inaccurate results may be obtained with (2) due to matching errors and improper assumptions of the scene. Therefore, photogrammetric approach is more suitable for accurate camera calibration. Furthermore, using a planar CB~\cite{zhang2000,zhang1998,plane_target_01,plane_target_02,plane_target_03} has advantages of flexibility and low cost compared with using a 3D target~\cite{3d_target_01,3d_target_02,3d_target_03}, it is widely adopted in traditional camera calibration, including the calibration for zoom-lens cameras. 

Specifically, images of a planar CB are firstly captured from at least two different viewpoints. Then, pairs of 3D points on the CB and their 2D counterparts on the image plane are identified and used to derive intrinsic and extrinsic parameters by minimizing reprojection errors. Since the optimization is based on algebraic distances, camera internal geometry may be estimated incorrectly, which may in turn deteriorate the estimation of the variation of camera parameters for zoom-lens camera calibration.

\noindent \textbf{Geometry-based Camera Calibration.} For more accurate camera calibration, a geometry-based method~\cite{geometry_01,geometry_02} is considered in this paper. Instead of completely relying on foregoing algebraic optimization mentioned above, an axis of geometric symmetry of each CB image, called principal line (PL), is firstly estimated with a closed-form solution. Subsequently, the PP, i.e., the center of symmetry, can be estimated as the intersection of at least two non-parallel PLs. Finally, the above information of PP and PLs can be used to estimate camera focal length and extrinsic parameters of CBs. It is shown in~\cite{geometry_01} that such approach is more accurate than the traditional calibration scheme mentioned above for synthetic data (and more robust for real data); therefore, it is adopted in this paper for the calibration of zoom-lens cameras. 

\subsection{Camera Calibration for Zoom-lens Cameras}
\label{sec_2.2}
Zoom-lens cameras play an important role in many applications for their superior ability to capture sizable images of objects, possibly with different depths in the scene. A look-up table is established in~\cite{zoom_depend_01} to record empirical results of several zoom/focus settings. However, such an approach is impractically inefficient as all possible zoom/focus settings need to be tested exhaustively. Therefore, a sampling-based test, followed by local interpolation approaches~\cite{interpolation_00,interpolation_01,interpolation_02,interpolation_03,interpolation_04}, is proposed in~\cite{zoom_depend_02}. In addition, global fitting methods~\cite{fitting_01,fitting_02,fitting_03,fitting_04} can also be adopted to systematically describe the variation of camera parameter. For example, best-fit of five or six estimated PPs and moving-least-squares-fit of PPs w.r.t. zoom/focus settings are further proposed in~\cite{zoom_depend_03} and~\cite{zoom_depend_04}, respectively. Ultimately, some algebraic approaches are developed in~\cite{zoom_depend_08,zoom_depend_06,zoom_depend_07} wherein all camera parameters are formulated as polynomials of focal length, or its reciprocal, before their coefficients are determined in the calibration process.

However, very different results are also obtained from other calibration schemes. For example, the PP associated with different focal lengths are assumed to have a fixed location in~\cite{zoom_depend_05} and estimated accordingly based on the simple concept of focus-of-expansion, while experimental analysis in~\cite{zoom_depend_03} actually suggests that randomness of PP locations exists for different settings of zoom and focus.

\subsection{Change of PP After Camera Rotation}
\label{sec_2.3}
On the other hand, to investigate possible variation of camera parameter due to different direction of gravity force, calibration results are also obtained by changing camera pose. Accordingly, several cameras are calibrated using target images captured at a fixed focal length, but with different camera poses in~\cite{gravity_02, gravity_01}. 
In~\cite{gravity_02}, based on the expectation that the effect of gravity on the lens may cause the optical axis to shift along the direction of gravity, and thus change the location of PP, calibration results are obtained for camera roll with two (0$^{\circ}$, 90$^{\circ}$) and four (0$^{\circ}$, 90$^{\circ}$, 180$^{\circ}$, $-90^{\circ}$) rotation angles.
However, the calibration results fall short of the foregoing expectation, as resultant changes in PP location appear to have practically no correlation with the rotation of the camera roll.

In~\cite{gravity_01}, similar experimental settings are performed for six identical cameras to investigate the effect of gravity on sensor-lens systems, for camera rolls with three rotation angles (0$^{\circ}$, 90$^{\circ}$, $-90^{\circ}$).
Although three clusters can be roughly identified for five (of six) of the resultant PP distributions, locations of these partially overlapped clusters, i.e., at top, middle, and bottom of central part of the image plane, cannot reasonably explain how the force of gravity, which has two orthogonal directions from the three camera rolls, affects the PP locations. Such results may be due to the use of only images which are obtained from different camera poses in their bundle adjustment process (four from one rotation angle and two for the other two rotation angles), leading to inaccurate calibration results. 

On the other hand, one possible reason of obtaining inconclusive results in Sec.~\ref{sec_2.2} and Sec.~\ref{sec_2.3} is the employment of traditional, algebra-based optimization schemes in the calibration process, which may not be accurate/robust enough for the derivation of small deviations of camera parameters due to changes in camera pose and focal length. To that end, the geometry-based camera calibration method~\cite{geometry_01} reviewed in Sec.~\ref{sec_2.1}, called Chuang’s method in this paper, is exploited together with the most used Zhang's method presented in~\cite{zhang2000}.

\section{Design of Experimental Procedure}
 \label{sec_3}
To address the issues mentioned above, several approaches are adopted/developed to enhance the performance of camera calibration under changes in lens location and the direction of gravity, which include:
\begin{enumerate}[(i)]
\item Maintaining best camera pan and tilt (PT) angles with respect to the orientation of the checkerboard (CB) pattern,
\item Ensuring image quality of a CB pattern according to~\cite{geometry_01}, using PLs,
\item Fixing camera pose for calibrating zoom-lens cameras,
\item Devising camera pose for investigating the effect of gravity,
\end{enumerate}
as will be elaborated in the following.

\subsection{Maintaining Best Camera PT Angles}
\label{sec_3.1}
In both Chuang’s method~\cite{geometry_01} and Zhang's method~\cite{zhang2000}, it is suggested that a dihedral angle between image plane and CB plane should be around 45$^{\circ}$ so that smaller errors in PP estimation can be achieved.\footnote{It is easy to see that the extreme cases of 0$^{\circ}$ and 90$^{\circ}$ will result in the absence of perspective effect and the diminishing image features, respectively.} Thus, a tilt/pan angle of around 45$^{\circ}$/0$^{\circ}$ with respect to the CB plane is maintained in the calibration.
Moreover, it is shown in~\cite{geometry_01} that a PP can be estimated as the intersection of a set of PLs, each obtained for a CB pattern. Therefore, it is suggested that the camera roll angles are evenly distributed in [0$^{\circ}$, 360$^{\circ}$). Accordingly, eight CB patterns roughly spaced by 45$^{\circ}$ in the above range will be maintained for the above tilt angle in all experiments considered in next section.

\subsection{Identifying and Improving CB Images of Poor Quality}
\label{sec_3.2}

\begin{figure}[!tb]
\begin{minipage}[t]{0.49\linewidth}
  \centering
  \vspace{0pt}
  \centerline{\includegraphics[width=4.2cm]{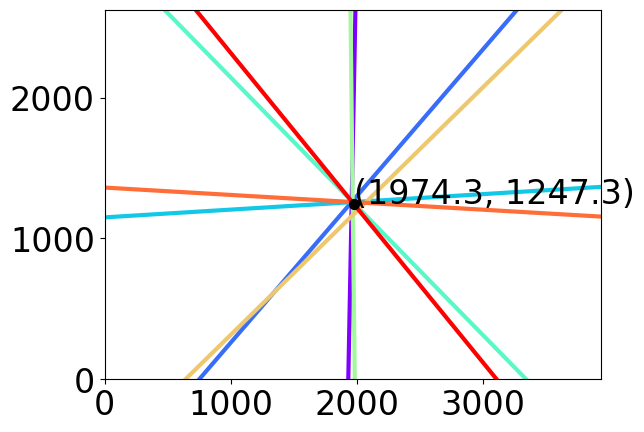}}
  \centerline{(a)}\medskip
\end{minipage}
\hspace*{\fill}
\begin{minipage}[t]{0.49\linewidth}
  \centering
  \vspace{0pt}
  \centerline{\includegraphics[width=4.2cm]{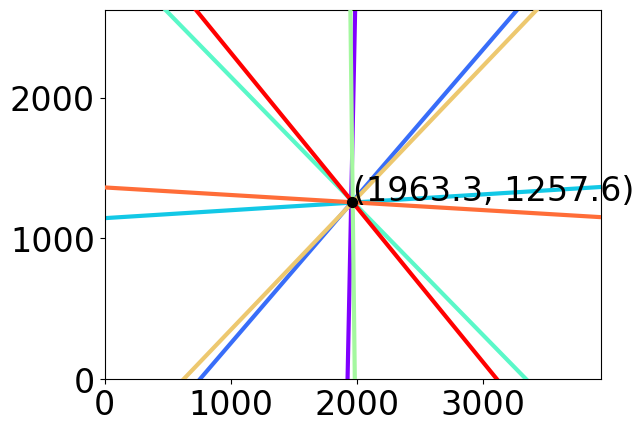}}
  \centerline{(b)}\medskip
\end{minipage}

\caption{(a) Eight PLs (one of them, the yellow one, obtained with extremely nonuniform illumination). (b) The problem in (a) resolved.}
\label{fig:example_abnormal_pl}

\end{figure}

Due to the geometric nature (intersection of PLs) of the PP estimation mentioned above, outliers of the PLs can be identified replaced easily, as suggested in~\cite{geometry_01}.
Fig.~\ref{fig:example_abnormal_pl} (a) shows a set of eight PLs with one of them obtained with extremely nonuniform illumination\footnote{Poor image quality may also result from image blurring due to camera motion or an out-of-focus CB} resulting in inaccurate corner detection of the CB image as well as an undesirable PL, i.e., the (yellow) PL which is deviated from a decent intersection of other PLs. Such a problem can then be resolved by (i) ensuring correct corner detection for this special CB image, e.g., with more involved image analysis techniques, as shown in Fig.~\ref{fig:example_abnormal_pl} (b), (ii) retaking the CB image with the same camera PT angle, or (iii) just removing that PL from the PP estimation. In our experiments, CB images improved by either (i) or (ii) will be used for both calibration schemes under consideration.

\subsection{Fixing Camera Pose for Calibrating Zoom-lens Cameras}
\label{sec_3.3}

\begin{figure}[!b]
\begin{minipage}[t]{0.3\linewidth}
  \centering
  \vspace{0pt}
  \centerline{\includegraphics[width=2.5cm]{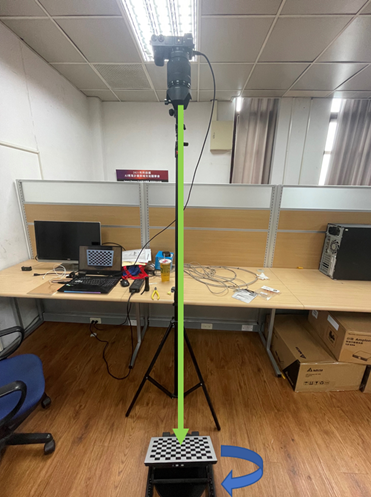}}
  \centerline{(a)}\medskip
\end{minipage}
\hspace*{\fill}
\begin{minipage}[t]{0.7\linewidth}
  \centering
  \vspace{0pt}
  \centerline{\includegraphics[width=6cm]{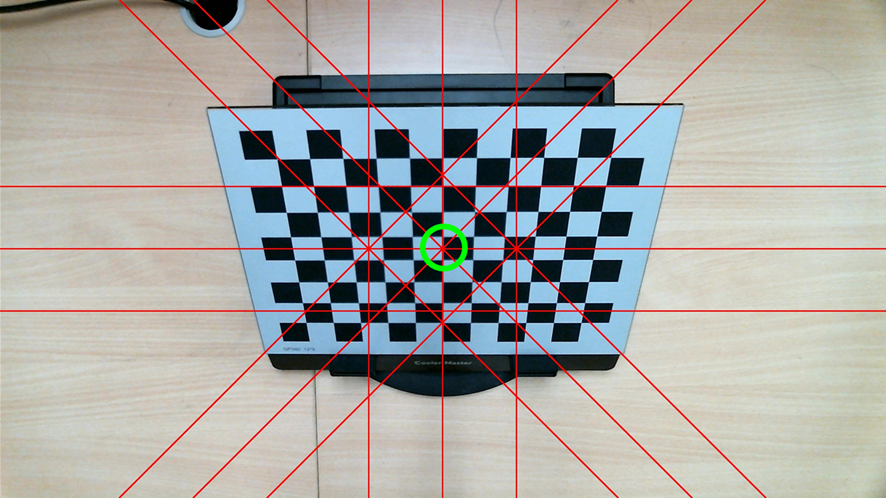}}
  \centerline{(b)}\medskip
\end{minipage}
   \caption{(a) A downward camera setting. (b) Auxiliary lines displayed on a viewing screen for image-CB alignment.}
\label{fig:camera_setting}
\end{figure}

While desirable PT angles are determined above for more accurate camera calibration, a stationary camera is required for more complicated situations in this paper, wherein camera changes in focal length and camera pose are both considered. Specifically, for the calibration of a zoom-lens camera, it is desirable to investigate the effect of focal length change alone, without such effect being interfered by the change in the direction of gravity due to camera pose change. Accordingly, for each focal length setting, the camera is fixed in a DOWN (downward) pose while the CB is placed on a horizontal rotation plate, with an elevation angle of 45$^{\circ}$ as shown in Fig.~\ref{fig:camera_setting} (a), before a set of eight CB images can be obtained for the calibration.\footnote{ In the experiments, we also aim the image center at the CB center of CB, as shown in Fig.~\ref{fig:camera_setting} (b), to reduce the effect of image distortion, if any.}

\begin{figure}[t]
\begin{center}
   \includegraphics[width=1.0\linewidth]{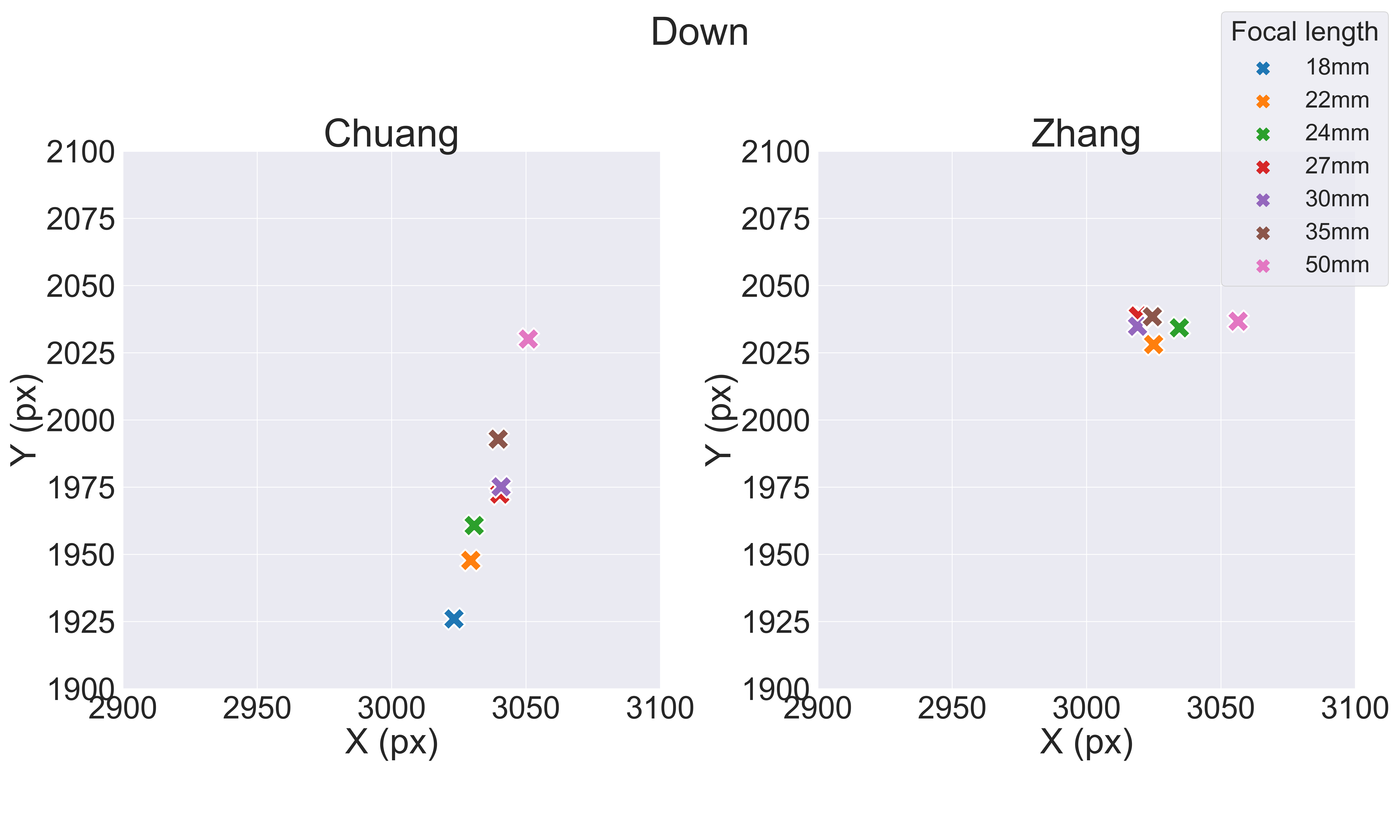}
   \centerline{~~~~(a)~~~~~~~~~~~~~~~~~~~~~~~~~~~~~~~~~(b)}\medskip
\end{center}
   \caption{PP shifts due to focal length changes, estimated with Chuang's method (left) and Zhang's method (right).}
\label{fig:cam2_fls_pps}
\end{figure}

Figs.~\ref{fig:cam2_fls_pps} (a) and (b) show loci of PP thus obtained by Chuang's and Zhang's methods for various setting of camera focal lengths, respectively. For Fig.~\ref{fig:cam2_fls_pps} (a), it is readily observable that the shift of PP in the NNE direction roughly increases monotonically with focal length. Such trend somewhat agree with PP identification results obtained from the optical experiments presented in~\cite{zoom_depend_optical}, which may result from the movement of imperfect lens or lens system, e.g., moving a tilted lens in a zoom-in action. However, Zhang’s results depicted in Fig.~\ref{fig:cam2_fls_pps} (b) are very different, i.e., there is no clear trend in the shift of PP for increasing focal length. More results for the calibration of zoom-lens cameras will be provided in the experiments.

\begin{table*}[b]
\small \addtolength{\tabcolsep}{-2.2pt}
\centering
\caption{
Cameras and their focal length samples used in experiments.
}
\label{tab:_exp_cam_setting}
\begin{threeparttable}
\begin{tabular}{|c|l|c|c|c|c|}
\hline
{} & Zoom Lens & Body & Image Resolution & Type* & Focal Length Samples\\
\hline
\emph{Cam} 1 & Sigma 18-50mm F2.8 EX DC MACRO HSM & Nikon D780 & 6048$\times$4024 & DSLR & 18, 22, 24, 27, 30, 35, 50 mm\\
\hline
\emph{Cam} 2 & Canon EF-S-18-50mm f/3.5-5.6 IS STM & Canon EOS 550D & 5184$\times$3456 & DSLR & 18, 23, 28, 30, 34, 39, 42 mm\\
\hline
\emph{Cam} 3 & Sony SEL2870 (28-70mm) & Sony $\alpha$6000 & 6000$\times$3376 & MILC & 33, 40, 44, 50, 55, 60, 65 mm\\
\hline
\emph{Cam} 4 & Sony SELP1650 (16-50mm) & Sony $\alpha$6000 & 6000$\times$3376 & MILC & 16, 21, 26, 33, 38, 45, 50 mm\\
\hline
\end{tabular}
\begin{tablenotes}  
\footnotesize
\item[*] DSLR: Digital Single Lens Reflex Camera
\item[*] MILC: Mirrorless Interchangeable Lens Camera
\end{tablenotes}

\end{threeparttable}
\end{table*}

\subsection{Devising Camera Pose for Investigating the Effect of Gravity}
\label{sec_3.4}
Besides the foregoing camera parameter variation due to focal length change, additional variation due to changes in the direction of gravity, similar to those considered in~\cite{gravity_01} and~\cite{gravity_02}, is also investigated in this paper.\footnote{While the former variation of camera parameters is mainly due to the contact force inside the lens system, the latter are due to changes in the non-contact force of gravity.} Accordingly, three additional camera poses, i.e., N (northward), W (westward), and E (eastward), are established to demonstrate possible PP shifts, in addition to those shown in Fig.~\ref{fig:cam2_fls_pps}, due to gravity.
Unlike the setup in~\cite{gravity_01} and~\cite{gravity_02}, these poses are obtained with small rotations (about 10$^{\circ}$) with respect to two axes of rotation, instead of one, for camera fixed in a generally downward pose.

\begin{figure}[tb]
\begin{center}
   \includegraphics[width=0.75\linewidth]{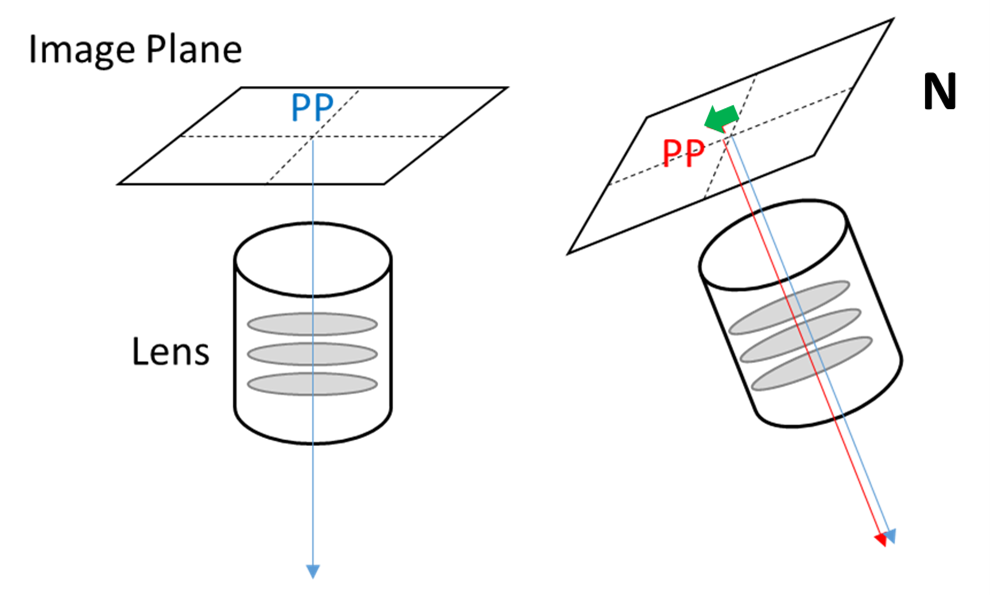}
\end{center}
   \caption{Side view of a camera with downward pose (left). A PP shift opposite to the direction of the gravity component in the image plane (green arrow) may occur in ICS when camera is rotated northward (right).}
\label{fig:gravity}
\end{figure}

Fig.~\ref{fig:gravity} shows the basic idea behind the experimental design. In particular, if the camera pose is changed from pointing downward (left) by rotating the camera northward (right), the internal lens configuration may experience additional gravitational force, resulting in a PP shift in the opposite direction (along the green arrow). Based on the same idea, three additional camera poses, i.e., N (northward), W (westward), and E (eastward), are established to demonstrate possible PP shifts, in addition to those shown in Fig.~\ref{fig:cam2_fls_pps}, due to gravity.

\begin{figure}[tb]
\begin{center}
   \includegraphics[width=1.0\linewidth]{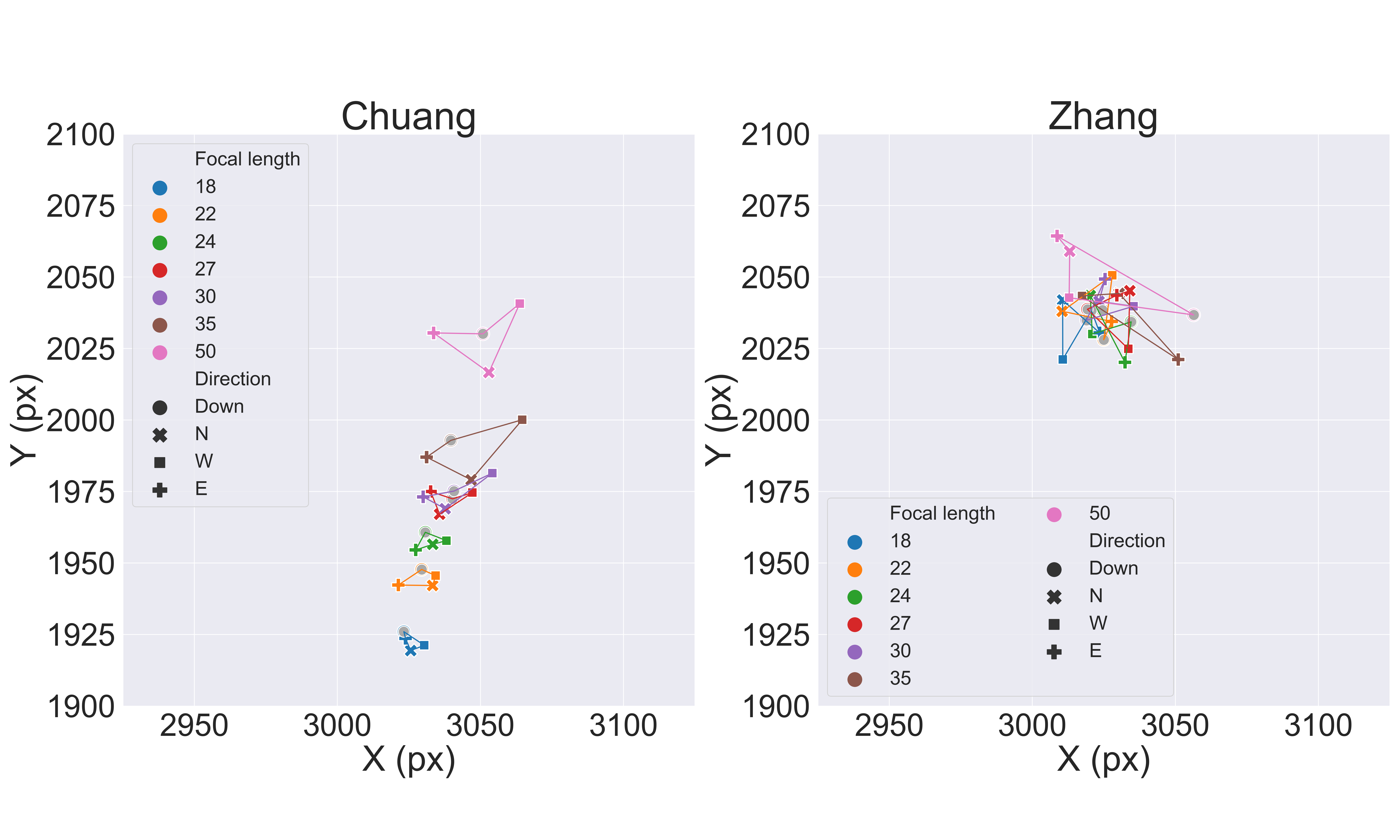}
\end{center}
   \caption{PP shifts due to (N, W, and E) camera pose changes, with PP shifts due to changes in camera focal length also displayed (in gray).}
\label{fig:cam2_camerapose_pps}
\end{figure}

Fig.~\ref{fig:cam2_camerapose_pps} shows the calibration results thus obtained. With the results shown in Fig.~\ref{fig:cam2_fls_pps} now depicted as gray circles, additional results obtained for N, W and E camera poses are represented by '$\times$', '$+$' and '${\mdblksquare}$', respectively. It is readily observable from Chuang’s result that additional minor (secondary) PP shifts, depicted with different colors for better visual examination, are added to the original (primary) PP shift due to focal length change. Moreover, directions of such minor PP shift are consistent to the (opposite) direction of gravity. However, the foregoing phenomena can not be observed from Zhang's results depicted in Fig.~\ref{fig:cam2_camerapose_pps} (right). Thus, more experimental results will be provided next to facilitate further investigation.

\section{Experiments}
In the previous section, experimental procedures are established for systematic evaluations of PP shift due to common physical changes in camera pose and focal length, based on two calibration schemes. However, inconsistent trends of PP shift are observed for the two schemes from preliminary results obtained with a selected camera, with clear trends obtained for both physical changes only from Chuang's method.

In this section, three additional zoom-lens cameras will be tested with calibration data established according to procedures developed in the previous section, as described in Sec.~\ref{sec_4.1}. Next, qualitative evaluations of PP shift due to changes in focal length and camera pose, similar to those presented earlier, will be provided in Sec.~\ref{sec_4.2} and Sec.~\ref{sec_4.3} for these cameras, respectively. Finally, quantitative evaluations based on cross-validation of intrinsic and extrinsic camera parameters obtained with different camera poses (for each focal length setting) will be provided in Sec.~\ref{sec_4.4}, in terms of reprojection error.

\subsection{Experimental Settings}
\label{sec_4.1}

\begin{figure}[!b]
\begin{minipage}[b]{1.0\linewidth}
  \centering
  \centerline{\includegraphics[width=6.5cm]{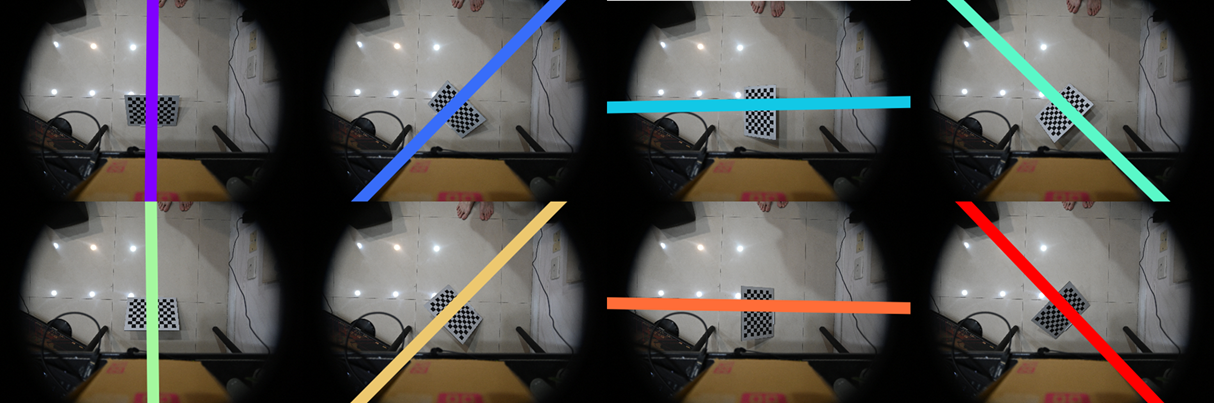}}
  \centerline{(a)}\medskip
\end{minipage}
\begin{minipage}[b]{1.0\linewidth}
  \centering
  \centerline{\includegraphics[width=6.5cm]{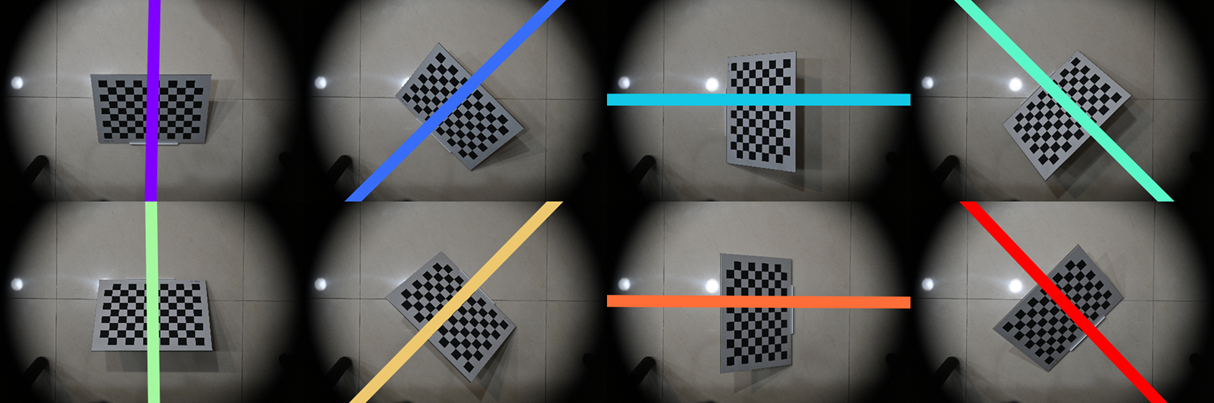}}
  \centerline{(b)}\medskip
\end{minipage}
\caption{Image dataset for the DOWN camera pose of \emph{Cam} 1,
with two focal length settings: (a) 22 mm and (b) 50 mm.}
\label{fig:dataset_zoom_in}
\end{figure}

\begin{figure}[!b]
\begin{minipage}[t]{0.45\linewidth}
  \centering
  \vspace{0pt}
  \centerline{\includegraphics[width=4.0cm]{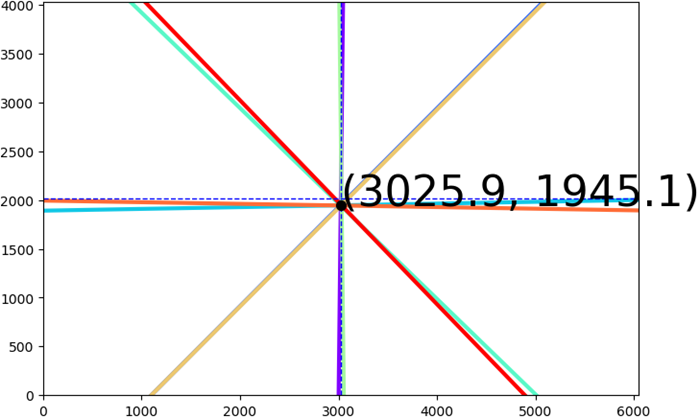}}
  \centerline{(a)}\medskip
\end{minipage}
\hspace*{\fill}
\begin{minipage}[t]{0.45\linewidth}
  \centering
  \vspace{0pt}
  \centerline{\includegraphics[width=4.0cm]{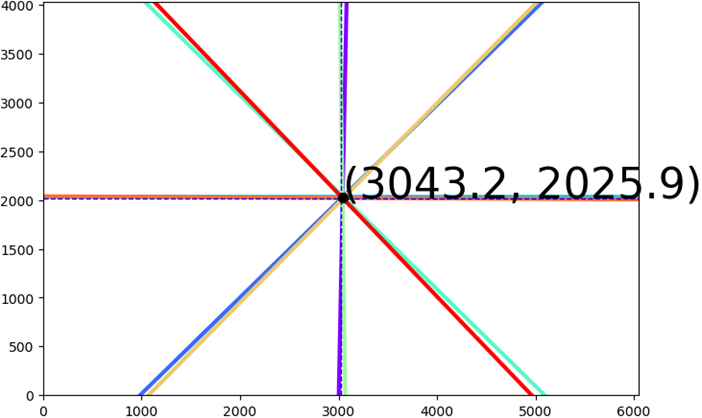}}
  \centerline{(b)}\medskip
\end{minipage}
\caption{ PP estimates obtained for (a) Fig.~\ref{fig:dataset_zoom_in} (a) and (b) Fig.~\ref{fig:dataset_zoom_in} (b).}
\label{fig:dataset_zoom_in_pp}
\end{figure}

\noindent \textbf{Camera Setting.} To extensively evaluate PP shift due to physical changes of camera, a number of cameras are employed in the experiments, as shown in Table~\ref{tab:_exp_cam_setting}. According to design of optical path, these cameras can be classified into digital single lens reflex (DSLR) camera or mirrorless interchangeable lens (MILC) camera. Furthermore, for DSLR, zoom lenses produced by both the original manufacturer and a sub-factory are selected. Effects of optical path design and assembly on the PP shift can thus be observed for more diverse camera systems.

\noindent \textbf{Calibration Datasets.} To estimate the shifted PP mentioned above, calibration data are prepared for four camera poses designed in Sec.~\ref{sec_3}, and seven focal lengths samples listed in Table~\ref{tab:_exp_cam_setting}, \footnote{These properly spaced samples are arbitrarily selected.} for each zoom-lens camera. Since eight CB images need to be captured using the alignment tool presented in Sec.~\ref{sec_3.1} in the calibration process for each PP estimate, 8$\times$7 CB images will be prepared for all focal length samples of each camera pose. Thus, the complete dataset obtained with the four cameras listed in Table~\ref{tab:_exp_cam_setting} will contain a total of 56$\times$4$\times$4 images.

\subsection{PP Distribution due to Focal Length Change}
\label{sec_4.2}

\begin{figure}[!b]
\begin{center}
   \includegraphics[width=1.0\linewidth]{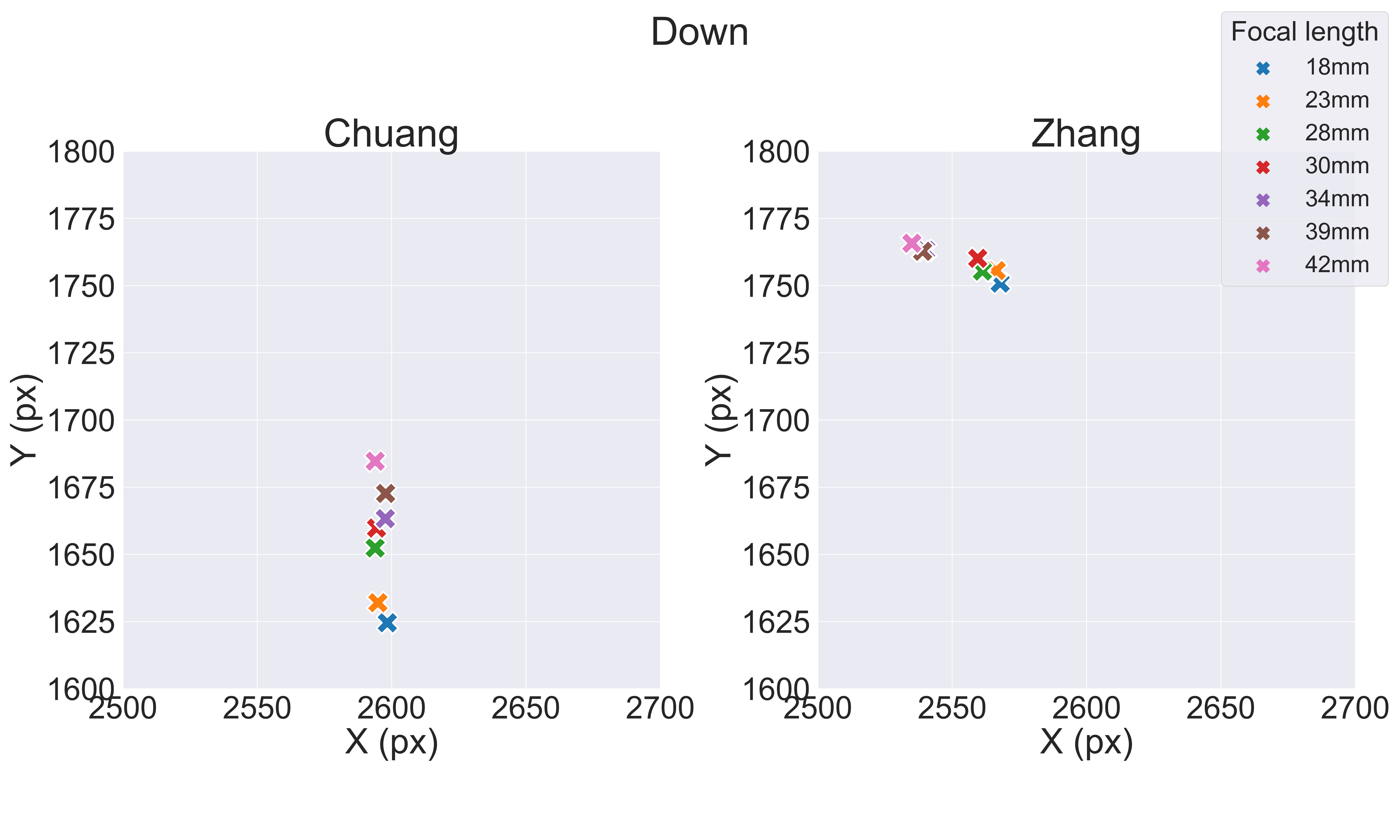}
\end{center}
   \caption{PP distributions of \emph{Cam} 2, with the DOWN camera pose, obtained with Chuang’s (left) and Zhang’s (right) methods.}
\label{fig:cam3_fls_pps}
\end{figure}

\begin{figure}[!b]
\begin{center}
   \includegraphics[width=1.0\linewidth]{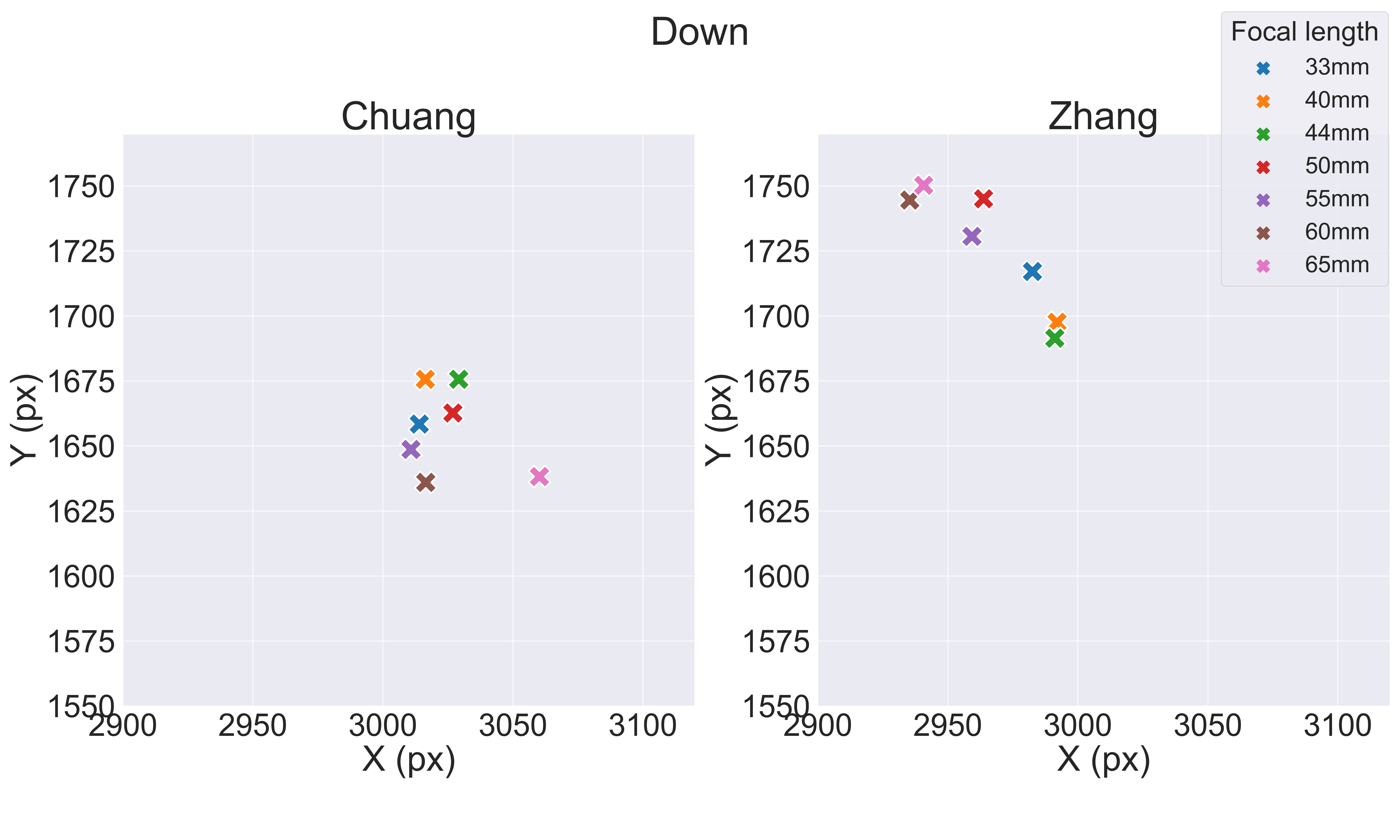}
\end{center}
   \caption{PP distributions of \emph{Cam} 3, with the DOWN camera pose, obtained with Chuang’s (left) and Zhang’s (right) methods.}
\label{fig:cam4_fls_pps}
\end{figure}

For the calibration of a camera with different focal length samples, CB images need to be collected.
In particular, Fig.~\ref{fig:dataset_zoom_in} shows two sets of images captured by \emph{Cam} 1 for two different focal length settings, i.e., 22 mm and 50 mm, together with the associated PLs. It is easy to see that different focal lengths correspond to CB images of different sizes. Each of the calibrated PPs, on the other hand, will need to be observed with a separate figure wherein the PP is supposely the intersection of all eight PLs, as shown in Fig.~\ref{fig:dataset_zoom_in_pp}. 

\begin{figure}[!tb]
\begin{center}
   \includegraphics[width=1.0\linewidth]{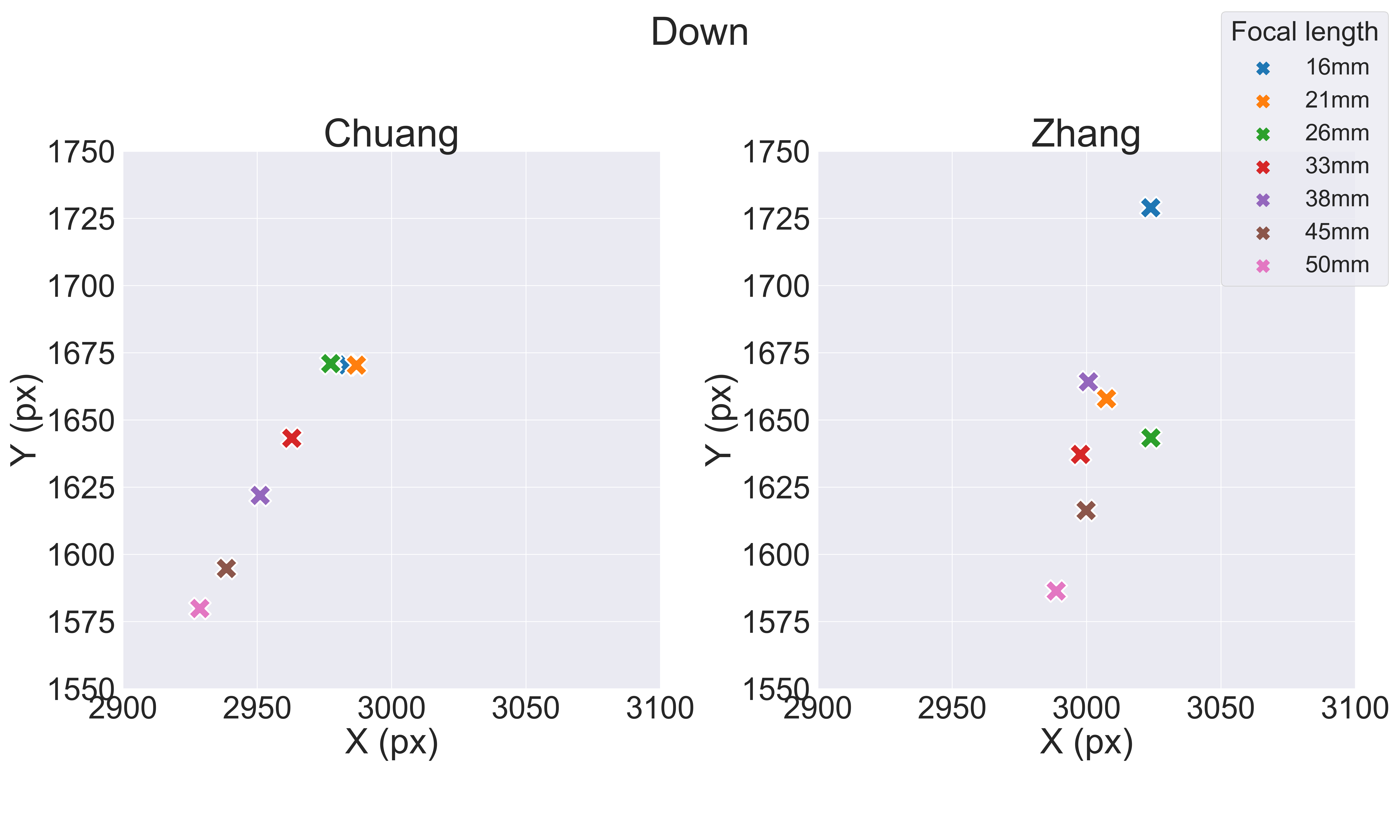}
\end{center}
   \caption{PP distributions of \emph{Cam} 4, with the DOWN camera pose, obtained with Chuang’s (left) and Zhang’s (right) methods.}
\label{fig:cam5_fls_pps}
\end{figure}

\begin{figure}[!b]
\begin{minipage}[b]{1.0\linewidth}
  \centering
  \centerline{\includegraphics[width=6.5cm]{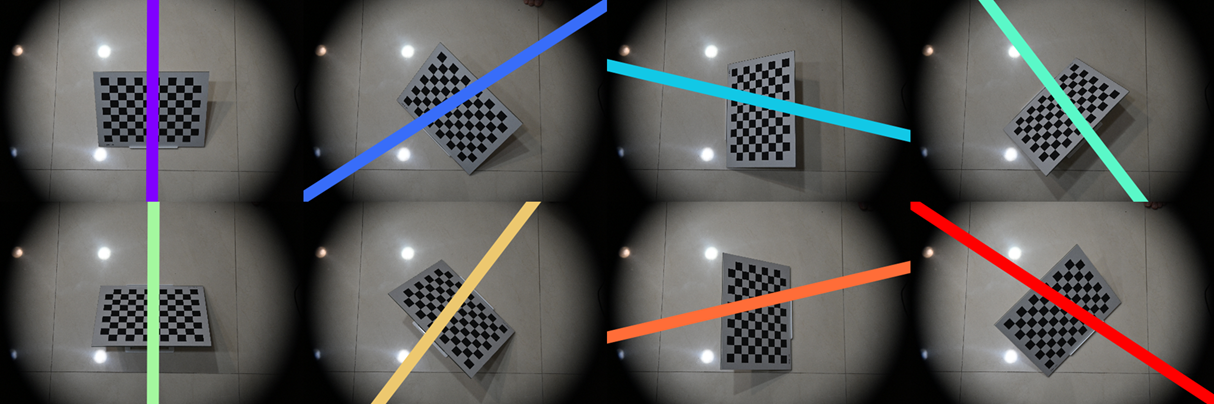}}
  \centerline{(a)}\medskip
\end{minipage}
\begin{minipage}[b]{1.0\linewidth}
  \centering
  \centerline{\includegraphics[width=6.5cm]{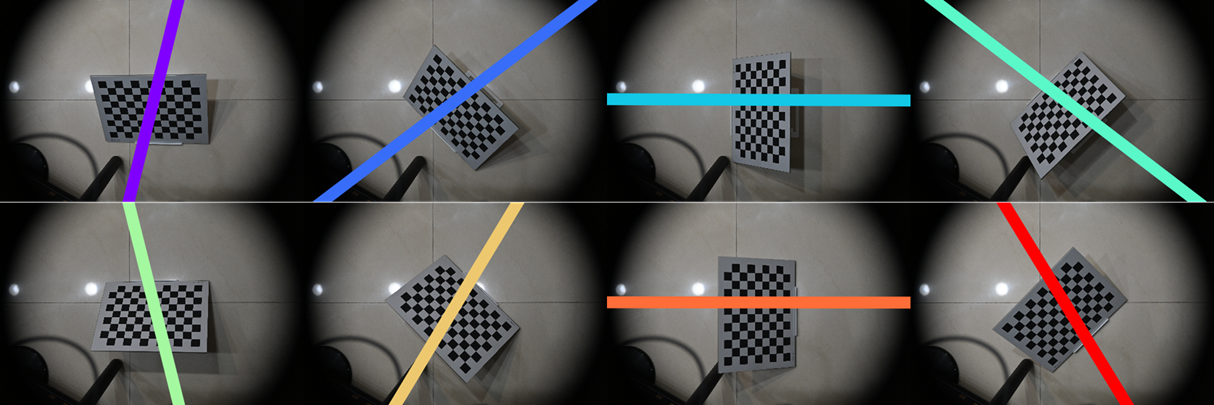}}
  \centerline{(b)}\medskip
\end{minipage}
\begin{minipage}[b]{1.0\linewidth}
  \centering
  \centerline{\includegraphics[width=6.5cm]{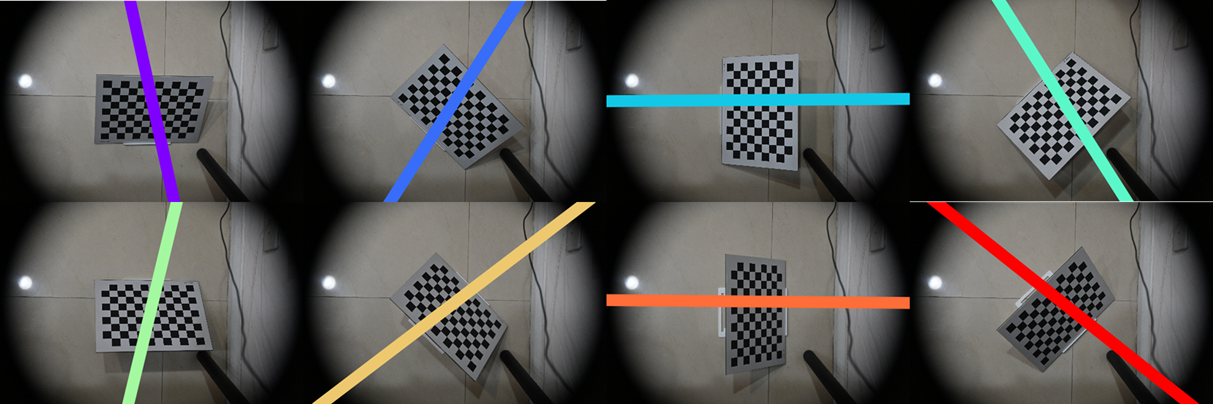}}
  \centerline{(c)}\medskip
\end{minipage}
\caption{Images obtained for \emph{Cam} 1, for three camera poses: (a) N, (b) W, and (c) E (see text).}
\label{fig:dataset_camera_pose}
\end{figure}

\begin{figure}[!b]
\begin{minipage}[t]{0.3\linewidth}
  \centering
  \vspace{0pt}
  \centerline{\includegraphics[width=3.0cm]{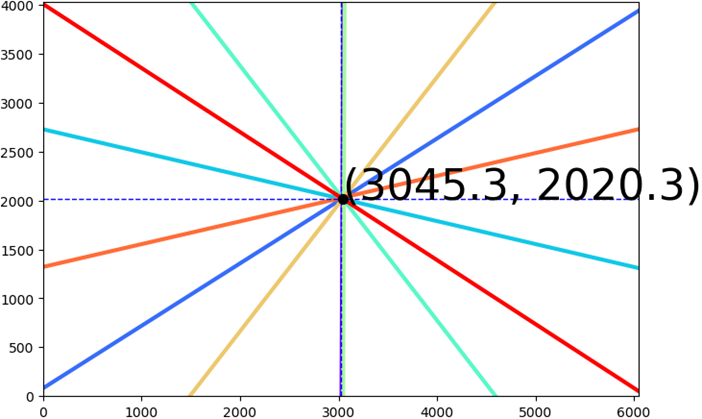}}
  \centerline{(a)}\medskip
\end{minipage}
\hspace*{\fill}
\begin{minipage}[t]{0.3\linewidth}
  \centering
  \vspace{0pt}
  \centerline{\includegraphics[width=3.0cm]{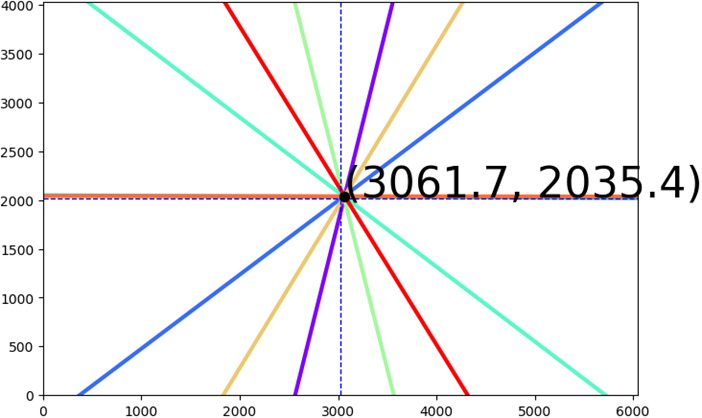}}
  \centerline{(b)}\medskip
\end{minipage}\hspace*{\fill}
\begin{minipage}[t]{0.3\linewidth}
  \centering
  \vspace{0pt}
  \centerline{\includegraphics[width=3.0cm]{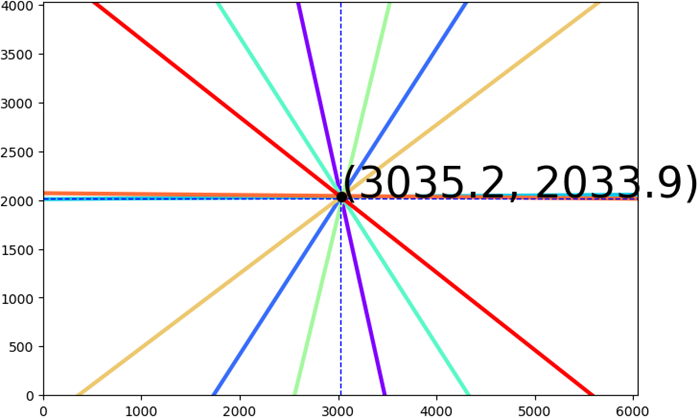}}
  \centerline{(c)}\medskip
\end{minipage}
\caption{PP estimates obtained for Figs.~\ref{fig:dataset_camera_pose} (a), (b), and (c).}
\label{fig:dataset_camera_pose_pp}
\end{figure}

Fig.~\ref{fig:cam2_fls_pps} and Figs.~\ref{fig:cam3_fls_pps}-\ref{fig:cam5_fls_pps} show the complete PP distribution obtained for \emph{Cam} 1 to \emph{Cam} 4, respectively, for the DOWN camera pose. While rather random PP shifts for increasing focal length are obtained by Zhang's method, major trends of the PP shift for each camera are readily observable from Chuang's results, which include: (i) it is roughly unidirectional, i.e., along NNE, N (with small magnitude and least fluctuations), roughly S (with small magnitude), and SSW direction, for \emph{Cam} 1 to \emph{Cam} 4, respectively, and (ii) the shift varies monotonically with the focal length.

In principle, Chuang's results seem to be more reasonable, and agree with results of the optical experiment presented in~\cite{zoom_depend_optical}. For (i), due to the rectilinear motion design of lens motor, lens in a zoom system is more likely moved in a specific direction, which may not be parallel to the optical axis of the camera. As for (ii), the aforementioned motion design is also likely to ensure that the PP is shifted monotonically during the zoom-in/zoom-out action.

\begin{figure}[!b]
\begin{center}
   \includegraphics[width=1.0\linewidth]{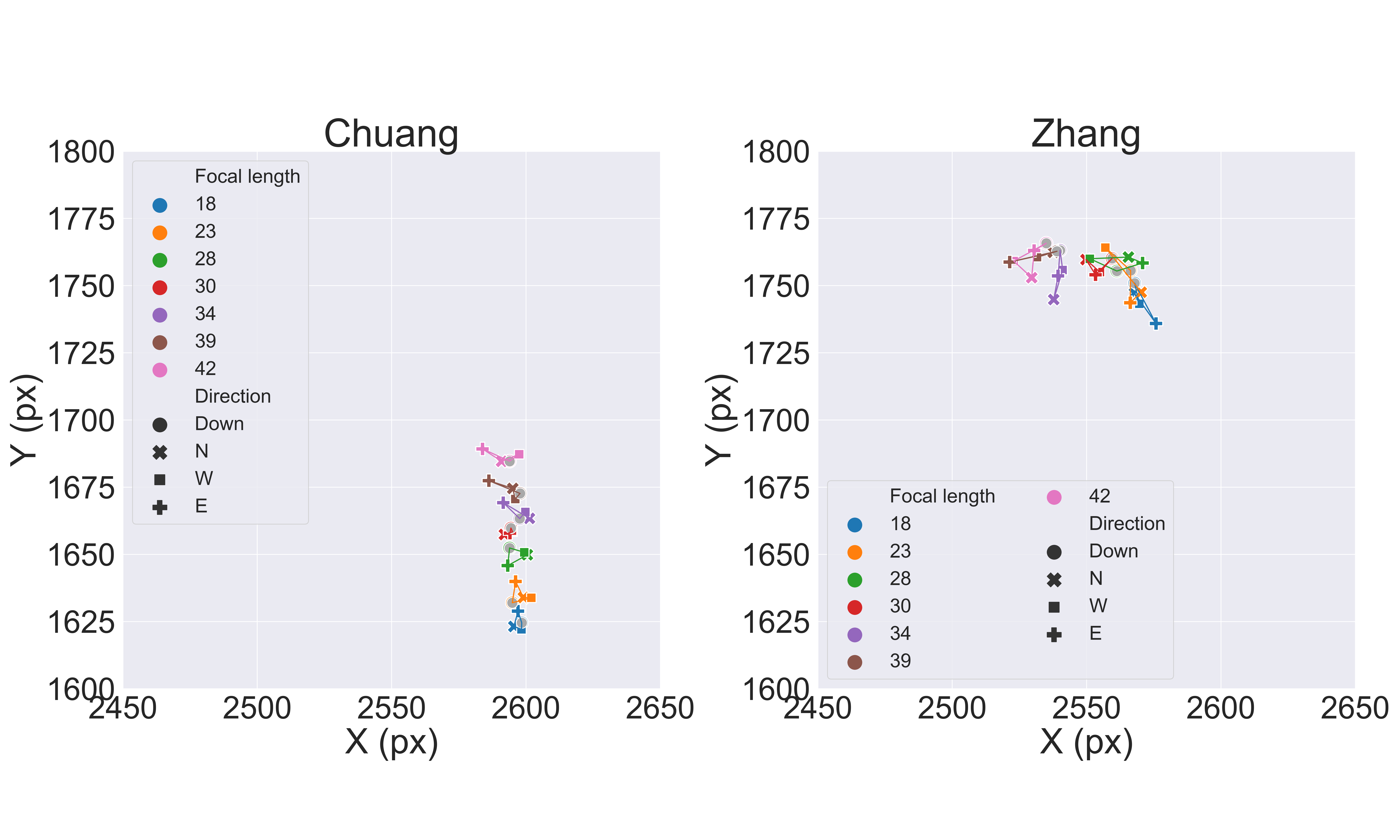}
\end{center}
   \caption{PP distribution of \emph{Cam} 2, with four camera poses tested for each focal length, obtained with Chuang’s (left) and Zhang’s (right) methods.}
\label{fig:cam3_camerapose_pps}
\end{figure}

\begin{figure}[!b]
\begin{center}
   \includegraphics[width=1.0\linewidth]{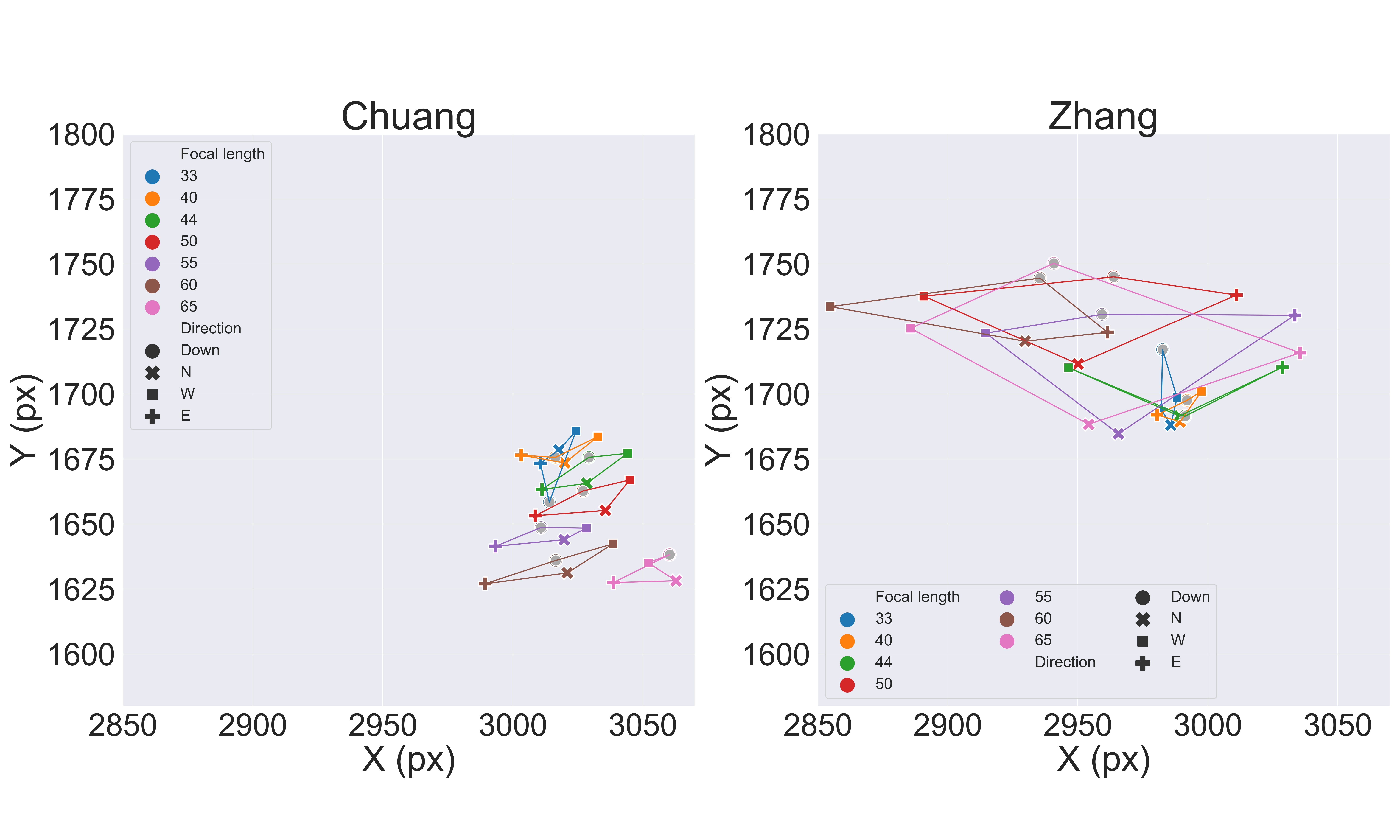}
\end{center}
   \caption{PP distribution of \emph{Cam} 3, with four camera poses tested for each focal length, obtained with Chuang’s (left) and Zhang’s (right) methods.}
\label{fig:cam4_camerapose_pps}
\end{figure}

\begin{figure}[!t]
\begin{center}
   \includegraphics[width=1.0\linewidth]{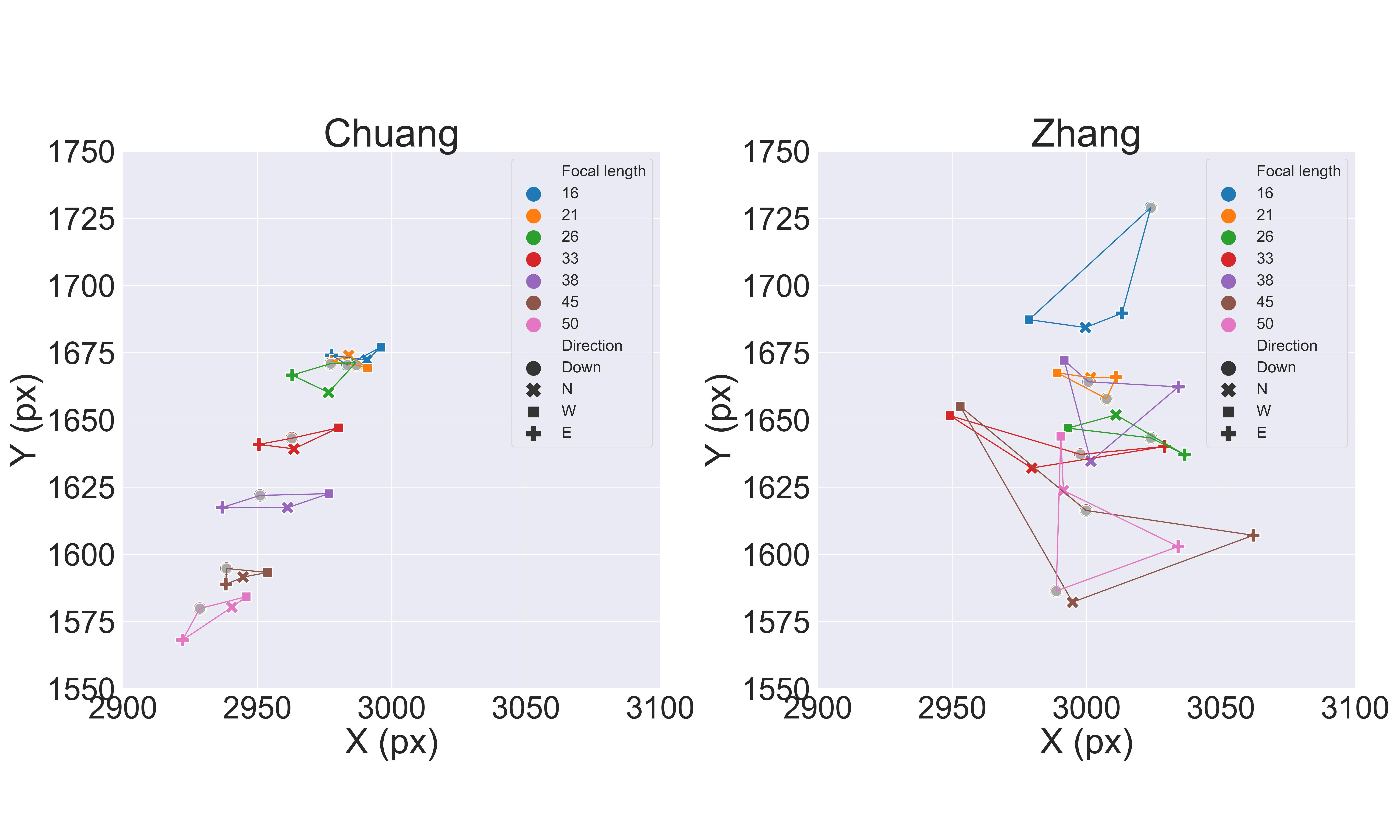}
\end{center}
   \caption{PP distribution of \emph{Cam} 4, with four camera poses tested for each focal length, obtained with Chuang’s (left) and Zhang’s (right) methods.}
\label{fig:cam5_camerapose_pps}
\end{figure}

\subsection{PP Distribution due to Camera Pose Change}
\label{sec_4.3}

In addition to the investigation of PP shifts due to contact force from possibly misaligned internal lens supporting system, shifts due to non-contact force, i.e., gravity, is further considered in this subsection, based on the experimental setup designed in Sec.~\ref{sec_3.4} for different camera poses, to further confirm the rationality of our explanation of PP deviation resulting from different runs of calibration.

Fig.~\ref{fig:dataset_camera_pose} shows image datasets similar to that shown in Fig.~\ref{fig:dataset_zoom_in} (b) but with three different camera poses. While the CB images have similar sizes as in Fig.~\ref{fig:dataset_zoom_in} (b), their orientations are not the same, causing variations in the direction of PL. Nonetheless, satisfactory estimation of PP can still be achieved by identifying the intersection of all PLs for each camera pose, as shown in Fig.~\ref{fig:dataset_camera_pose_pp}, wherein the PP deviation due to camera pose change can be observed. 

Fig.~\ref{fig:cam2_camerapose_pps} and Figs.~\ref{fig:cam3_camerapose_pps}-\ref{fig:cam5_camerapose_pps} show the complete PP distribution obtained for different poses (and for each focal length) of \emph{Cam} 1 to \emph{Cam} 4, respectively. While rather random PP shifts for four camera poses (and for increasing focal length) are obtained by Zhang's method, minor (secondary) trends of the PP shift (on top of the major (primary) one due to focal length change) can be observable from Chuang's results, which include: (i) it is along the direction roughly opposite to the gravitational force and (ii) the shift is most sideway to that due to focal length change, or having larger vector component perpendicular to the trajectory of gray circles in the figures.

Again, Chuang's results for the foregoing PP shift seem to be more reasonable, in addition to the fact the environmental force actually results in the same trends for different cameras. Specifically, the gravity results in consistent triangular patterns of {$\times$, $+$, ${\mdblksquare}$} for (i), with "$\times$" almost always below the line segment connecting "$+$" and "${\mdblksquare}$" and "$+$" ("${\mdblksquare}$") always located on the right (left). As for (ii), the rationality is established upon a reasonable assumption that the contact force from the internal lens configuration has already pushed the system toward the direction of the primary PP shift discussed in the previous subsection, leaving a space mostly for a movement in the perpendicular directions for the non-contact force, or gravity, and resulting in triangles elongated in these directions.

\section{Conclusion}
In this paper, a novel calibration procedure is proposed to identify the major and minor PP deviations due to changes in focal length and camera pose. Based on Chuang’s calibration method~\cite{geometry_01}, major (unidirectional and monotonic) trends of PP shift can be obtained for focal length change, which are different for different types of camera, possibly due to different internal lens configurations, while minor (mostly sideway and consistent to the direction of gravity for all cameras) PP shifts can be obtained for camera pose change. In addition to these qualitative analyses, quantitative evaluations based on reprojection errors are developed with the cross-validation of intrinsic and extrinsic camera parameters obtained with different camera poses (for each focal length setting), and ultimately confirm the validity of our investigation on variations of camera parameter and their causes, for both zoom-lens cameras and cameras with fixed focal length.

\section*{Acknowledgment}
This research was supported by the Ministry of Science and Technology, Taiwan, R.O.C. under contract NSTC 110-2221-E-A49-083-MY2.

\ifCLASSOPTIONcaptionsoff
  \newpage
\fi

\bibliographystyle{unsrt}
\bibliography{main}

%

\vspace{-5 mm}
\begin{IEEEbiography}[{\includegraphics[width=1in,height=1.25in,clip,keepaspectratio]{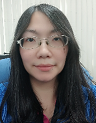}}]{Hsin-Yi Chen}
received the BS degree in Applied Mathematics from National Chung Hsing University, Taiwan, in 2005, MS degree in Computer Science and Information Engineering from National Central University, Taiwan, in 2007. She then served as a deputy engineer in Industrial Technology Research Institute mainly focusing on optimization researches from 2007 to 2012. From 2012 to 2015, she served as an engineer and mainly focused on algorithm and system development in AOI (automatic optical inspection) field. From 2015 to 2019, she served as a senior engineer and focused on AI (artificial intelligence) applications which include smart factory and interactive system of transparent display. Since 2018, she joins the full-time doctoral program in Department of Computer Science of National Chiao Tung University.
\end{IEEEbiography}

\begin{IEEEbiography}[{\includegraphics[width=1in,height=1.25in,clip,keepaspectratio]{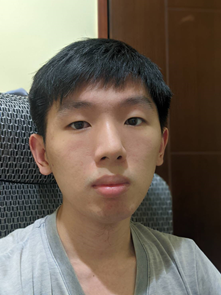}}]{Chuan-Kai Fu}
received the BS degree
in Computer Science from National Sun Yat Sen University, Taiwan, in 2019, MS degree in Computer Science and Information Engineering from National Yang Ming Chiao Tung University, Taiwan, in 2022.
\end{IEEEbiography}

\begin{IEEEbiography}[{\includegraphics[width=1in,height=1.25in,clip,keepaspectratio]{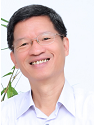}}]{Jen-Hui Chuang}
received the BS degree in electrical engineering from National Taiwan University in 1980, the MS and Ph.D. degrees, both in electrical and computer engineering, from University of California at Santa Barbara and Urbana-Champaign University of Illinois at Urbana-Champaign in 1883 and 1991, respectively. Since then he has been on the faculty of the Department of Computer and Information Science of National Chiao Tung University (NCTU). From 2004 to 2005, he was the Chairman of the Department of Computer and Information Science of NCTU. From 2006 to 2007, he was the Associate Dean of College of Computer Science. From 2017 to 2020, he served as the Dean of College of Computer Science of NCTU. Dr. Chuang’s research interests include signal and image processing, computer vision and pattern recognition, robotics, and potential-based 3-D modeling. He served as an associate editor of Signal Processing from 2005 to 2010 and has been an associate editor of Journal of Information and Science and Engineering since 2009. Dr. Chuang was the President of Chinese Image Processing and Pattern Recognition Society from 2015 to 2016 and served as the governing board member of International Association of Pattern Recognition from 2013 to 2016. Currently, he serves as the governing board member of IEEE Taipei Chapter. Dr. Chuang is a senior member of IEEE.
\end{IEEEbiography}

\end{document}